# A Study imbalance handling by various data sampling methods in binary classification

Mohamed Hamama[1]

**Abstract** The purpose of this research report is to present the our learning curve and the exposure to the Machine Learning life cycle, with the use of a Kaggle binary classification data set and taking to explore various techniques from pre-processing to the final optimization and model evaluation, also we highlight on the data imbalance issue and we discuss the different methods of handling that imbalance on the data level by over-sampling and under sampling not only to reach a balanced class representation but to improve the overall performance. This work also opens some gaps for future work.

## 1 Introduction

The selected data set (*Kaggle, Rain in Australia [1]*) is holds 10 years of daily weather measurements and observations from different locations in Australia.
Several weather stations located across Australia gathers various weather measurements like temperature, humidity and observations like cloud covers and those metrics also include the amount of measured rain in mm as (Rainfall). The following list shows the column fields and their missing value counts.

1. **Date**            : The date of the observation (nulls =0)
2. **Location**        : The weather station location name (nulls =0)
3. **MinTemp**         : Minimum temperature °C (nulls =637)
4. **MaxTemp**         : Maximum temperature in °C (nulls =322)
5. **Rainfall**        : The amount of the day rainfall (mm) (nulls =1406)
6. **Evaporation**     : The evaporation in (mm) (nulls =60843)
7. **Sunshine**        : The number of hours of bright sunshine (nulls =67816)
8. **WindGustDir**     : The Direction of strongest wind gust (nulls =9330)
9. **WindGustSpeed**   : The Speed of strongest wind gust (km/hr) (nulls =9270)
10. **WindDir9am**     : The Direction of wind at 9 AM (nulls =10013)
11. **WindDir3pm**     : The Direction of wind at 3 PM (nulls =3778)

1 Cork Institute of Technology, Department of Computer Science, Mohamed.hamama@mycit.ie (R00194200)

12. **WindSpeed9am**   : The Speed of wind at 9 AM (km/hr) (nulls =1348)
13. **WindSpeed3pm**   : The Speed of wind at 3 PM (km/hr) (nulls =2630)
14. **Humidity9am**    : The Humidity in percentage at 9 AM (nulls =1774)
15. **Humidity3pm**    : The Humidity in percentage at 3 PM (nulls =3610)
16. **Pressure9am**    : The Atmospheric pressure at 9 AM (hpa) (nulls =14014)
17. **Pressure3pm**    : The Atmospheric pressure at 3 PM (hpa) (nulls =13981)
18. **Cloud9am**       : The fraction sky coverage in eighths at 9AM (oktas) (nulls =53657)
19. **Cloud3pm**       : The fraction sky coverage in eighths at 3PM (oktas) (nulls =57094)
20. **Temp9am**        : The Temperature at 9AM (°C) (nulls =904)
21. **Temp3pm**        : The Temperature at 9AM (°C)) (nulls =2726)
22. **RainToday**      : Boolean: 1 if precipitation (mm) exceeds 1 mm (nulls =1406)
23. **RISK_MM**        : The amount of next day mm (nulls =0)
24. **RainTomorrow**   : Boolean: 1 if the precipitation (mm) of next day RISK_MM > 1 mm

The dataset is interesting topic and it is widely used which the prediction of the rain expectancy in future and exploring this ML topic is worth exploration. The dataset creator has stated that a new variable (RISK_MM) which is the Rainfall for tomorrow was used to generate the Target classification variable RainTomorrow based on Yes if RISK_MM > 1 mm, hence this column should be removed to avoid leaking information to the model predictions [2]. A separate study for RISK_MM as target regression value would be interesting for a future work.

This dataset will help us to evaluate the impact of the imbalance in the binary classification problem and it will provide good baseline to explore and evaluate various techniques to overcome the problem and improve the precision in predicting the positive class and our evaluation metric will be the F1 score. The dataset shows imbalance in the class representation with the positive label (Yes) has 31877 samples out of 142193 total records and the negative label (No) has 110316 samples.

We will start by establishing the performance baseline and then we will explore the various techniques and their impact.

## 2    Research

A challenging fact that a real-life data usually an imbalanced data, which means a class maybe represented by large number of samples while the other class has lower representation in the data set this will lead to predictions biased to the majority class. *This happens when once class represents a circumscribed concept, and the other class represents the counterpart of that concept* as per [4]. This could be found in medical records datasets, anomaly detection which may be highly imbalanced. In Machine Learning world learning algorithms assume that the training data are balanced, and the provided samples contain equal numbers of the target class labels which is not the real-world scenario. For the example chosen for the purpose of this paper the Rain in Australia dataset which suffers from 77:23 imbalance. The aim of this research report is to study and evaluate some techniques that are widely used to handle the imbalance problem in machine learning. An interesting finding by the authors of [5] that not only class imbalance would hinder the performance of the classification algorithms but also related to data overlapping degree between the classes, hence the pre-processing stage may have a great impact on how the models perform.

Due to the commonality of the class imbalance in the classification problems the accuracy might not be a good representation for the model performance this should be evaluated by the confusion matrix [6] were we could extract representative numbers for the model performance like the Precision and the Recall and their harmonic mean the F1-score shall be more representative for the model performance. A binary classification where the positive class is the target we use the previous metrics calculated for the positive class and then the precision is called specificity and the recall is called sensitivity, and since the F1 score which is the harmonic mean for both Precision and Recall [3].

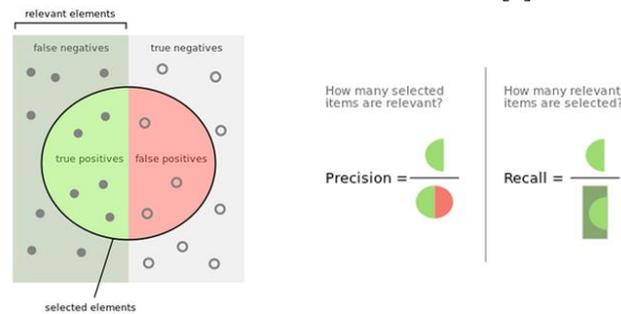

**Figure 1 Precision and Recall**

$$F_1 = 2 * \frac{precision * recall}{precision + recall}$$

**Figure 2 F1 the Harmonic Mean for Recall and Precision**

The Imbalanced dataset may be handled either by handling the imbalance on *the data level to change the distribution of the imbalanced data sets* [11] to improve better classification. The second category is by algorithm level by modifying the algorithms to avoid the imbalance problems for example using different kernel function for SMV classifiers or using a Balanced Bagging

Classifier to use balanced bag samples. In this research paper we will discuss only the first category.

## 2.1 Imbalance Handling on the Data level

In this research report we will explore various techniques of imbalance handling, One those techniques is based on altering the sample data used for training the machine learning model this either by under sampling the majority class or oversampling the minority class until both classes balance, and recently some techniques combines of over sampling and under sampling may be used to improve the performance of the learning model.

In under sampling the target is to select a sample of examples representing the majority class that are equal to the number of minority class examples, which means we are eliminating data points from majority class until every class is equally representation. On other hand the over sample is achieved by generating replicating samples randomly to increase the representation of the minority class. Those concepts in some situations they may outperform the complex techniques like SMOTE and Tomek links

### 2.1.1 Synthetic Minority Over Sampling Technique (SMOTE)

Rather than generating random samples a synthetic technique is used to select random points between a two neighboring data points in the minority class the aim is to avoid overfitting that may occur by replicating minority class samples while using the random over sampling [7].

### 2.1.2 Tomek Links

It is a technique where it establishes a virtual links between the two nearest neighbors from opposite class while each example is the nearest to the other. Then Tomek Link either be used as an under sampling by deleting examples for the majority class or as a *data cleaning technique by deleting the examples of both classes considering that they are either noise or borderlines* [4] [8].

### 2.1.3 SMOTE + Tomek Links

This was first used by the authors of [1] to apply Tomek Links cleaning method to the SMOTE over-sampled training set this is handle the deformation in the class clusters that may happen due to oversampling which may lead to overfitting. And to achieve this when Tomek Links method applied to the oversampled data it does not remove only the majority class data but by eliminating examples from both classes to achieve a well-defined cluster. The application of the method illustrated in Figure 3 [4].

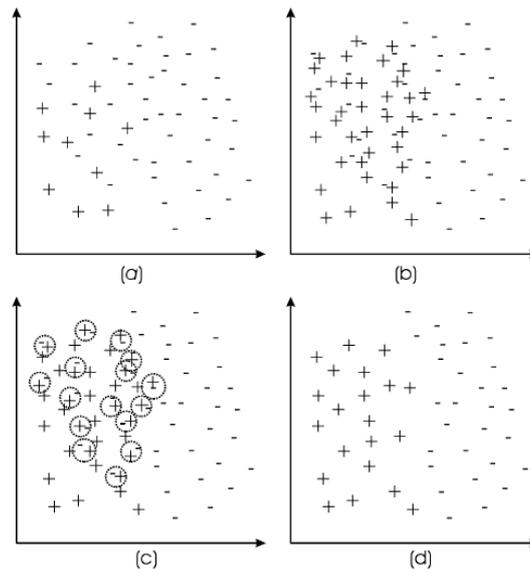

**Figure 3** *Balancing a data set: original data set (a); oversampled data set (b); Tomek links identification (c); and borderline and noise examples removal (d)* **[4].**

### 2.1.4 SMOTE + ENN

A Similar approach to the use of SMOTE with Tomek Links is to use SMOTE combined with *Wilson's Edited Nearest Neighbor Rule (ENN)* [9] who proposed the removal of any example with a label differs from the class of the nearest two of its three neighbor or the nearest k-1 of its k-NN. When SMOTE+ENN is applied it tend to remove more samples than SMOTE + Tomek Links[4].

### 2.1.5 SMOTE BorderLine

Because of the impotency of the borderline examples contribution to the classification the authors of [11] have proposed modifications to SMOTE method, that rather than selecting a random minority examples we target only borderline examples and their neighbors to synthetically generated interpolated examples. The experimentations in [11] showed improved True positive rates and F-values than SMOTE and random sampling. It would be interesting if we explore the impact of this method on classification performance.

**2.1.6 Adaptive Synthetic Sampling Approach (ADASYN)**

The *essential idea of ADASYN is to use weighted distribution for different minority class examples according to their level of difficulty of learning* [12] to generate more samples for the examples that are harder to learn and this aims to improve the learning of the classifiers. The weight is the density distribution which is the normalized ratio of the number neighbors that belong to the majority class to the total number of neighbors.

# 3 Methodology

## *3.1 Part 1. Establishing Baseline*

**3.1.1 Preprocessing**

- Dropped Features the following features:
    a. **RISK MM** is a regression variable adjacent to the target classification variable
    b. **Sunshine** contains more than 47% of null values
    c. **Evaporation** contains more than 42% of null values
    d. **Cloud3pm** contains more than 40% of null values
    e. **Cloud3pm** contains more than 37% of null values

- A new engineered feature **"Month"** is extracted from Date, the Day will be ignored as Month only correlate to seasonal changes in the climate and the Year is time series variable also will be ignored

- Categorical features/Labels handling:
    a. **RainTomorrow** using Ordinal Encoder Yes/No to 1/0

- b. **Location, WindowGustDir, WindowDir9am, WindowDir3pm, RainToday** On-Hot Encoder using Pandas get_dummies.
- Imputation of Null values, using Simple imputation to replace null values with Average/Mean of the feature or the Iterative Imputer (MICE) is used but in the model selection phase I have used the Simple Imputer only.
- Outlier Removals using IsolationForest with contamination value of 1% which was not used in model selection and then evaluated in the following experimentations
- Normalization all numerical columns in the dataset was normalized using MinMaxScaler.

### 3.1.1 Model Selection

An initial range of basic models from different categories with their default parameters, a printout of F1 score for the positive class "Yes" as the goal of the classification problem to predict if it is going to RainTomorrow based to the validation/testing class frequencies, the models were fitted without any imbalance resampling techniques.
The program parameters used where the following:

```
balancer_sel = 0    # 1: SMOTE , 2: SMOTETomek, 3: SMOTEENN, 4: ADASYN Else No re-Balance
dim_reducer_sel = 0    # Enable Feature Selection 1:SelectKBest(Based on kBest value)  2: RFECV 3: SelectFwe 4 : PCA
outliers_remover_sel = 0
contamination = 0.01
```

The MLPClassifer, RandomForestClassifier, and LogisticRegression were the best scorers on the initial run, but considering the long time consumed fitting and optimizing MLP Classifier, the choice for hyperparameter optimization and the experimentation will go to Random Forest and Logistic Regression.

### 3.1.1 Hyper Parameter Optimization

Using GridSearchCV to optimize the selected models (Random Forest Classifier, Logistic Regression Classifier)
The optimization process returned the following tuning of the models:

```
models_for_experiment = [("RandomForestClassifier", RandomForestClassifier(max_features='sqrt',
                                                                          min_samples_split=4,
                                                                          n_estimators=300,
                                                                          min_samples_leaf=2,
                                                                          n_jobs=-1)),
                         ("LogisticRegression", LogisticRegression(C=100,
                                                                   solver='saga',
                                                                   max_iter=2000,
                                                                   tol=1e-3,
                                                                   n_jobs=-1))]
```

## *3.2 Part 2. Basic Experimentation*

I have done lots of experimentations in order to improve the pre-processing and feature selections by having the following:
- Removal of the Location and its dummy columns and I found that it reduces the classification's F-value.
- Following the recommendations of [13] that IsolationForest is performing better when used after the IterativeImputer (MICE) so I deviated from the Assignment recommendations by not starting with outlier detection. But In the comparison with the use of Interquartile outlier detection and capping their values then using the SimpleImputer to fill missing values with the mean I found that the improvement of the f_value does not exceed .01 (1%)
- I have experimented with various feature selection algorithms: Like SelectKBest(chi2), Recursive Feature Elimination RFECV and SelectFwe and I experimented with PCA. With the following mapping with dim_reducer_sel parameter, the Select by Family wise-error alogrethim performed very well by selecting 86 features and then I will continue with it.

```
1: SelectKBest(chi2, k=kBest)
2: RFECV(estimator=LogisticRegression(solver="liblinear"), step=1, cv=5, scoring='f1', n_jobs=-1)
3: RFECV(estimator=RandomForestClassifier(), step=1, cv=10, scoring='f1', n_jobs=-1)
4: SelectFwe(chi2)
5: PCA(n_components=kBest)
```

## *3.3 Part 3. Research*

A set of selected Samplers (Balancer) from imblearn API have been utilized for the purpose of the evaluation of their impact on the performance of RandomForestClassifier and LogisticRegression Classifier after establishing the baseline and performing the experimental tunings those Samplers could be found in the following code snippet

A Study imbalance handling by various data sampling methods in binary classification

```
balancers = [("SMOTE", SMOTE(n_jobs=-1)),
             ("Combined SMOTE & Tomek", SMOTETomek(n_jobs=-1)),
             ("Combined SMOTE & ENN", SMOTEENN(n_jobs=-1)),
             ("Adaptive Synthetic (ADASYN)", ADASYN(n_jobs=-1)),
             ("Borderline SMOTE", BorderlineSMOTE(n_jobs=-1)),
             ("RandomOverSampler", RandomOverSampler()),
             ("RandomUnderSampler", RandomUnderSampler()),
             ]
# The Value of balancer_sel parameter changes the active sampler:
1: SMOTE ,2: SMOTETomek, 3: SMOTEENN, 4: ADASYN, 5: BorderlineSMOTE,
6: RandomOverSampler, 7: RandomUnderSampler 0: No Balance Resampling
```

The goal is to change perform the evaluation using a 10 fold cross validation for 80% of the data, also the program will evaluate the performance against the all data set.

## 4 Evaluation

On the first run of the model selection phase the The MLPClassifer, RandomForestClassifier, and LogisticRegression were the best scorers on the initial run as per the below table, but I have decided to discard the use of Multilayer Perceptron Classifier for the computational cost and time consumed. And I started investigating the pre-processing techniques.

| Model | F1-score |
|---|---|
| **MLPClassifier()** | 0.6044 |
| **RandomForestClassifier()** | 0.5974 |
| **LogisticRegression()** | 0.5650 |
| DecisionTreeClassifier() | 0.5238 |
| QuadraticDiscriminantAnalysis() | 0.4897 |
| GaussianNB() | 0.4734 |
| PassiveAggressiveClassifier () | 0.3512 |
| KNeighborsClassifier() | 0.3511 |

Table 1 – Initial Model configurations score of first run

**Outlier Detection and Missing Value handling:** The use of Iterative Imputer and Isolation Forest slightly improved the f_value instead of the use of Interquartile method for capping the outliers and then Simple Imputer. The results while using Borderline SMOTE

| Model | F1-score with IterativeImputer , IsolationForest and BorderlineSMOTE | F1-score with Interquartile, SimpleImputer and BorderlineSMOTE |
|---|---|---|
| | | |

| RandomForestClassifier() | 0.6460 | 0.6342 |
|---|---|---|
| LogisticRegression() | 0.6164 | 0.5988 |

**Table 2 Outlier Detection and Missing Value experimentation f1_value**

I decided to continue with Interquartile, SimpleImputer for the computational burden implied by the Iterative Imputer and the IsloationForest.

**Feature Selection and Dimensionality Reduction:** 10 Folds cross validation run with an 80% as training/validation set and 20% as a Test set, Borderline SMOTE was used to resample the training sets . The results are displayed in Table 3 Using RFECV with RandomFoestClassifier estimator returned 10 Features their ordered top 5 are: "MaxTemp", "Rainfall", "WindGustSpeed", "WindSpeed3pm", "Humidity9am". While using RFECV with Logistic regression estimator returned 54 features their ordered top 5 are: "Rainfall", "WindGustSpeed", "Humidity9am", "Humidity3pm", "Pressure9am".

| Model | 0 : No Selection (117 Features) | 1: SelectKBest (50 Features) | 2: RFECV (10 Features) RandomForest | 3: RFECV (54 Features) Logistic Regression | 4: SelectFwe (87 Features) | 5: PCA (50 Components / Features) |
|---|---|---|---|---|---|---|
| RandomForestClassifier()-CV | 0.6535 | 0.6379 | 0.6307 | 0.6429 | 0.6497 | 0.587 |
| LogisticRegression()-CV | 0.6147 | 0.5959 | 0.5889 | 0.6006 | 0.6080 | 0.558 |
| RandomForestClassifier()-Test | 0.652 | 0.634 | 0.629 | 0.636 | 0.647 | 0.586 |
| LogisticRegression()-Test | 0.608 | 0.592 | 0.589 | 0.589 | 0.607 | 0.559 |

**Table 3 Feature selection experimentation f1_value**

Family Wise Error alogorith returned 84 features and performed a slightly better than Logistic Regression Selection by RFECV and overall better than other methods the feature selection methods this may be due to smaller number of k in SelectKBest or n_components in the PCA, I have experimented values of 10, 20 and 30 and the f_value is improving with the increased number of features it would be interesting if I increase that number to 87 feature and compare the results with the other methods. The Principal Component Analysis PCA dimensionality reduction showed the worst performance. The selection final evaluation for the research topic will follow with the Family Wise-Error SelectFwe as a feature selection.

**Imbalanced Data Sampling Methods:** 10 Folds cross validation run with an 80% as training/validation set and 20% as a Test set, Borderline SMOTE was used to resample the training sets. The results are displayed in Table 3

| Model | 0: Imbalance | 1: SMOTE | 2: SMOTETomek | 3: SMOTEENN | 4: ADASYN | 5: BorderlineSMOTE | 6: RandomOverSampling | 7: RandomUnderSampling |
|---|---|---|---|---|---|---|---|---|
| RandomForestClassifier() | **0.5974** | **0.6516** | 0.6491 | **0.6224** | 0.6496 | 0.6493 | 0.6486 | 0.6334 |

| -10 Fold Cross Validation | | | | | | | | |
|---|---|---|---|---|---|---|---|---|
| RandomForestClassifier() - 20% Test Set | **0.602** | 0.648 | 0.647 | **0.621** | **0.652** | 0.647 | 0.649 | 0.637 |
| LogisticRegression() -10 Fold Cross Validation | **0.5824** | 0.6145 | 0.6131 | **0.5860** | 0.6055 | 0.6081 | **0.6147** | 0.6115 |
| LogisticRegression() - 20% Test Set | **0.585** | 0.607 | 0.610 | **0.585** | 0.610 | 0.609 | 0.612 | 0.622 |

**Table 4 Resampling Techniques f1_value best performers in green and worst in red**

The Random forest classifier has shown very good improvement with almost all the techniques with improvements up to 8-9% in f-value with SMOTE and ADASYN then SMOTETomek and BorderlineSMOTE those results are the averages of the 10 fold cross validation and even with 80% Training set and 20% training set the f-value was slightly lower than cross validation average. The SMOTEEN method was the lowest improvement event less than using random methods but also it showed improvement 3-4% in f-value in both cross validation and test set. The Random under sampling showed good results 5-6% improvement it would be interesting if I have sometime to check it for overfitting, the most interesting observation was the Random over sampling performance was showing results between 7.8%-8.5% which was similar and may out performing some of the complex techniques which is also concluded in [4].

With the Logistic Regression model, the Random methods have out performed all the other complex methods with an improvement 4.6%-5.5% for random over sampling between the test set and the cross validation, the under sampling showed even more improvement 5%-6.3% this may because we are having positive minority class having more than 32K samples which represent around 23% of the total samples. The SMOTE, BorderlineSMOTE, ADASYN, SMOTETomek has showed similar results that are ranging between 3.7%-5% between training/test set and the cross validation, but the interesting fact that SMOTEENN has performed the worst and the Trianing/Test it has 0% improvement in comparable to the imbalanced dataset.

A note that the used F1 score (f_value) is the harmonic mean for the positive class Precision and recall while the f_value for the negative majority calss may exceed 80% with some models as we have explored in the initial baseline preparations. Also, there was models like Naïve Bayes (GuassianNB) it is performance was very sensitive for any variations in the feature selection and sampling method, it may be explored

## 5 Conclusion

In this research report we explored the Machine Learning life cycle, from data exploration to the final model selection and optimization, also we analyzed some data level methods to over-sample or under-sample the imbalanced data set in order to improve the learning of the minority positive class, Although there are too many complex methods that aims for improving the classification performance either by balancing the data or even to interfere with its distribution over the feature space the random methods are still competing for improving classification performance and even for their computational speed.
Also, it worth mentioning that there were models like Naïve Bayes (GuassianNB) where its performance was very sensitive for any variations in the feature selection and sampling method, it may be explored deeply in a future work.

A an interesting future investigation in feature selection could be conducted by the use of domain knowledge, where some of the features are recorded in a specific time like wind, pressure, humidity and temperature at 9am and 3pm those could be grouped into two subgroups and a clustering algorithm could be used in order to generate a cluster distribution that represent label that reflects the Weather Status at 9AM and at 3PM, other features like location could be a part of each group the final labels Weather9AM and Weather3PM could replace all the grouped features and then investigated for effectiveness in this data set and this exactly would be similar to the work presented by [14] and similar to the concept discussed in [15].

**Note:** Potential improvements to the methodology of this research is that currently in this research project, I have applied many preprocessing steps on the full data set for example fitting the Scaler on the full data set rather than using only the training data and separately predict the testing set scales, this also should be implemented on the cross-validation level prevent validation data statistical information to leak to the training set, this fact was learned to late in this project work. Another potential improvement may be achieved by different techniques in feature selection by extensive study in their correlation an *interesting work in Kaggle notebooks [16] and [17] using this data set is work to consider also those Kaggle notebooks helped a lot in achieving this work.*

## References [2]

---

[2] Note that the references cited by the report the work and the code examples.

A Study imbalance handling by various data sampling methods in binary classification